\documentclass[conference]{IEEEtran}

\usepackage{cite}
\usepackage{amsmath,amssymb,amsfonts}
\usepackage{algorithmic}
\usepackage{graphicx}
\usepackage{textcomp}
\usepackage{xcolor}

\usepackage{booktabs}

\def\BibTeX{{\rm B\kern-.05em{\sc i\kern-.025em b}\kern-.08em
    T\kern-.1667em\lower.7ex\hbox{E}\kern-.125emX}}
    
\begin{document}

\title{Low-Field Magnetic Resonance Image Quality Enhancement using Undersampled k-Space and Out-of-Distribution Generalisation}

\author{
\IEEEauthorblockN{Daniel Tweneboah Anyimadu$^1$, Mohammed M. Abdelsamea$^1$, Ahmed Karam~Eldaly$^{1,2}$}
\IEEEauthorblockA{
$^1$Department of Computer Science, University of Exeter, Exeter, United Kingdom\\
$^2$UCL Hawkes Institute, Department of Computer Science, University College London, London, United Kingdom\\
\{da536, m.abdelsamea, a.karam-eldaly\}@exeter.ac.uk
}
}

% \author{\IEEEauthorblockN{1\textsuperscript{st} Daniel Tweneboah Anyimadu}
% \IEEEauthorblockA{\textit{Department of Computer Science
% } \\
% \textit{University of Exeter}\\
% Exeter, United Kingdom \\
% da536@exeter.ac.uk}
% \and
% \IEEEauthorblockN{2\textsuperscript{nd} Mohammed M. Abdelsamea}
% \IEEEauthorblockA{\textit{Department of Computer Science} \\
% \textit{University of Exeter}\\
% Exeter, United Kingdom \\
% M.Abdelsamea@exeter.ac.uk}
% \and
% \IEEEauthorblockN{3\textsuperscript{rd} Ahmed Karam Eldaly}
% \IEEEauthorblockA{\textit{Department of Computer Science} \\
% \textit{University of Exeter}\\
% Exeter, United Kingdom \\
% A.Karam-Eldaly@exeter.ac.uk}
% }

\maketitle

\begin{abstract}
Low-field magnetic resonance imaging (MRI) offers affordable access to diagnostic imaging but faces challenges such as prolonged acquisition times and reduced image quality. Although accelerated imaging via k-space undersampling helps reduce scan time, image quality enhancement methods often rely on spatial-domain postprocessing. Deep learning achieved state- of-the-art results in both domains. However, most models are trained and evaluated using in-distribution (InD) data, creating a significant gap in understanding model performance when tested using out-of-distribution (OOD) data. To address these issues, we propose a novel framework that reconstructs high-field-like MR images directly from undersampled low-field MRI k-space, quantifies the impact of reduced sampling, and evaluates the generalisability of the model using OOD. Our approach utilises a k-space dual channel U-Net to jointly process the real and imaginary components of undersampled k-space, restoring missing frequency content, and incorporates an ensemble strategy to generate uncertainty maps. Experiments on low-field brain MRI demonstrate that our k-space-driven image quality enhancement outperforms the counterpart spatial-domain and other state-of-the-art baselines, achieving image quality comparable to full high-field k-space acquisitions using OOD data. To the best of our knowledge, this work is among the first to combine low-field MR image reconstruction, quality enhancement using undersampled k-space, and uncertainty quantification within a unified framework.
\end{abstract}


% \vspace{-0.15cm}
\section{Introduction}
\label{sec:intro}
% \vspace{-0.15cm}
Low-field MRI (LF-MRI; $<1$ T) offers an affordable alternative to high-field MRI \cite{arnold2023low}, but faces two significant challenges: prolonged scan times and intrinsically lower signal-to-noise ratios (SNR), both of which compromise image quality \cite{ayde2025mri, arnold2023low}. Moreover, while MRI data are typically acquired in the frequency domain (k-space), fully sampling k-space is time-consuming; practical protocols often rely on undersampling to accelerate acquisition, which leads to an ill-posed inverse problem that further degrades image quality. To mitigate these challenges, recent techniques such as compressed sensing and deep learning have improved reconstruction from undersampled k-space data \cite{eldaly2024bayesian, safari2025advancing, souza2019hybrid, kiryu2023clinical, lyu2023m4raw, Eldaly2026ICASSP}. However, image quality enhancement is typically performed in the spatial domain, either through super-resolution (SR) or image quality transfer (IQT) \cite{alexander2014image, alexander2017image, lin2023low, kim20233d, eldaly2024alternative, iglesias2023synthsr, Daniel2026ISBI, Tien2026ISBI}, which learns mappings between high-quality reference scans and their low-quality counterparts to recover missing information in LF-MRI scans.

Although effective, these methods operate separately, with IQT/SR techniques applied to spatial domain images reconstructed using k-space \cite{alexander2014image, alexander2017image, lin2023low, kim20233d, eldaly2024alternative, kimberly2023brain, islam2023improving}. This separation fails to address both prolonged scan times and poor resolution within a unified framework. As a result, valuable raw phase and magnitude information is lost, which is vital for preserving fine anatomical details, especially in LF-MRI, where SNR limitations are more pronounced. Moreover, the integration of uncertainty quantification (UQ) \cite{lambert2024trustworthy}, which complements reconstruction by providing uncertainty maps that highlight regions of reduced model confidence and can support cautious interpretation and downstream analyses such as region-of-interest selection, remains underexplored in the context of LF-MRI, particularly under k-space undersampling conditions \cite{zou2023review, lambert2022trustworthy}. These issues highlight the need for a novel, unified framework that accelerates image acquisition, enhances image quality while simultaneously quantifying reconstruction uncertainty.

On the other hand, while data-driven models are typically trained on closed world scenarios, real world data often differs. Existing methods tend to focus on in-distribution (InD) data, leaving a gap in understanding how models perform when tested using OOD data, particularly in LF-MRI, where SNR limitations are more pronounced \cite{eldaly2024alternative}. Thus, the main contributions of this work are as follows. (1) We propose a novel framework that reconstructs high-field MR like images from undersampled LF-MR k-space measurements, while quantifying the effect of the reduced samples using OOD data. To the best of our knowledge, this is among the first attempts in the literature to combine LF-MR image reconstruction, quality enhancement, and uncertainty quantification in a unified framework. (2) We propose a U-Net architecture that jointly processes real and imaginary channels of k-space, enabling accurate restoration of missing frequency information. (3) We adopt a cross-validation-based ensemble strategy to generate confidence maps, which highlight high-error regions and enhance the reliability of reconstructions. (4) We demonstrate that our k-space-driven IQT approach achieves superior performance over the
% counterpart spatial-domain method, and yields
spatial-domain counterparts and existing methods, yielding reconstructions from undersampled OOD data that are comparable in quality to full high-field k-space acquisitions.

% \vspace{-0.2cm}
\section{Method}
\label{sec:method}
% \vspace{-0.2cm}
{In this section, we present an end-to-end framework designed to address LF-MRI reconstruction, image quality enhancement, and UQ using undersampled k-space data. Our framework employs a k-space dual channel U-Net architecture, as illustrated in Figure \ref{fig:U-Net}, which processes both the real and imaginary components of complex-valued k-space, represented as two real-valued channels and stored as floating-point tensors, to generate high-field-like k-space outputs from undersampled acquisitions. The architecture uses convolutional operations and nonlinear activations to extract hierarchical features from the two-channel k-space input, while interleaved pooling layers in the encoder progressively enhance salient features and reduce spatial dimensionality. Following the standard U-Net design, the decoder reconstructs high-field–like outputs from the bottleneck representation using transposed convolutions and skip connections, supporting the recovery of fine-scale details. The final output preserves the real and imaginary k-space channels, allowing the model to exploit complementary magnitude and phase information while preserving the underlying complex-valued structure of the data during learning. Accordingly, the two-channel formulation can be viewed as approximating complex-valued convolution through coupled operations on the real and imaginary components, expressed as:
\begin{equation}
\quad (\mathbf{W} \ast \mathbf{y}) = (\mathbf{W_r} \ast \mathbf{y_r} - \mathbf{W_i} \ast \mathbf{y_i}) + i(\mathbf{W_r} \ast \mathbf{y_i} + \mathbf{W_i} \ast \mathbf{y_r}),
\end{equation}
where \(\mathbf{W_r}, \mathbf{W_i}\) are the real and imaginary components of the convolutional weights, respectively, and \(\mathbf{y_r}, \mathbf{y_i}\) are the corresponding observations. This coupled formulation helps preserve both magnitude and phase structure throughout the reconstruction process.

\begin{figure}%[htbp]
\centering
\includegraphics[width=0.995\columnwidth]{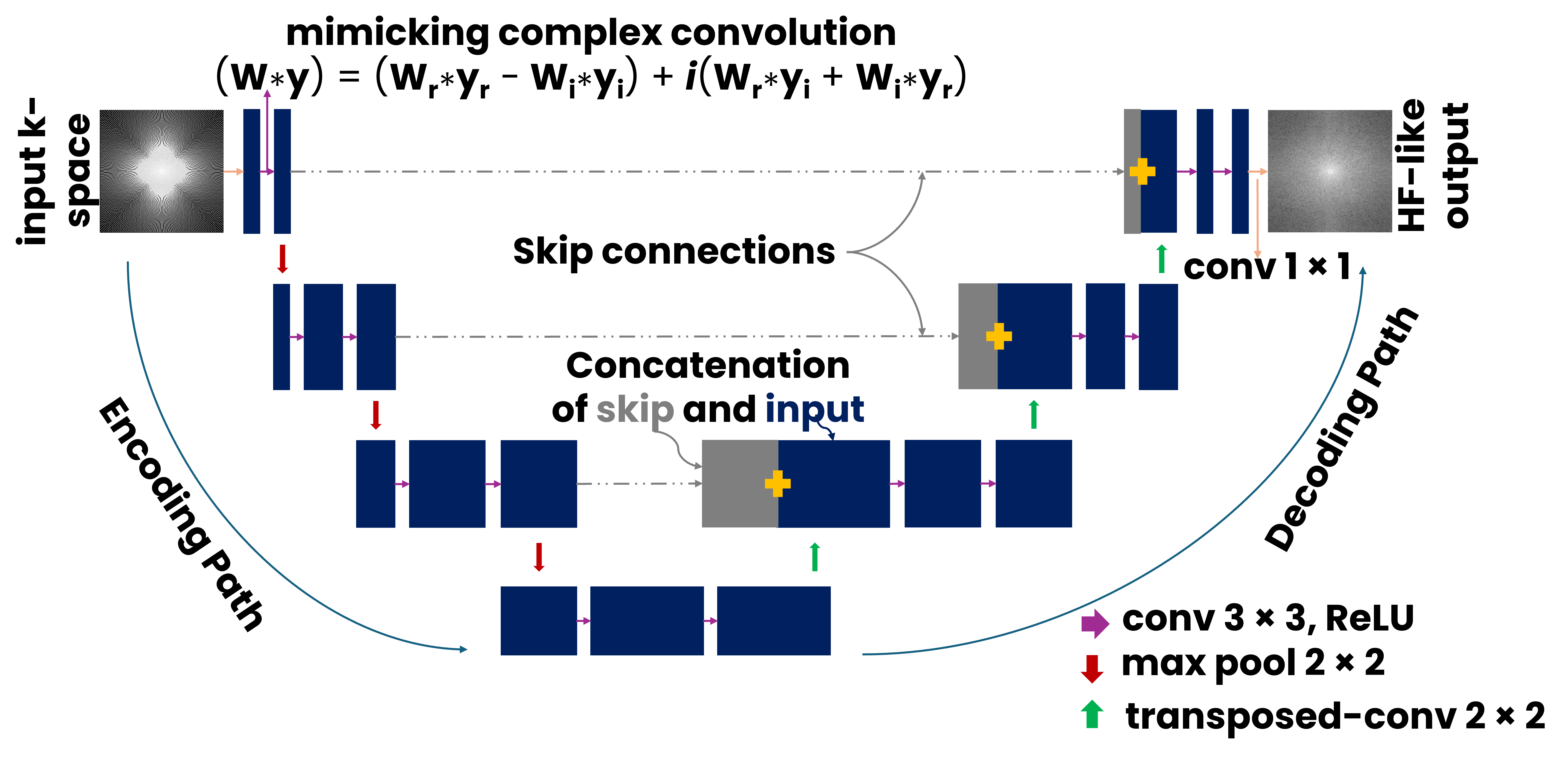}
% \vspace{-0.5cm}
\caption{K-space dual channel U-Net for low-field MRI k-space reconstruction, quality enhancement and uncertainty quantification.}
\label{fig:U-Net}
% \vspace{-0.4cm}
\end{figure}

To achieve high-field-like reconstruction and image enhancement from undersampled data, we train our k-space dual channel U-Net model using paired input-output examples. These examples are created by transforming high-field-MR images into synthetic LF counterparts, applying randomised tissue-specific contrasts to capture the variability observed in real LF scans \cite{lin2019deep, lin2023low}. The corresponding k-space data are then undersampled at varying acceleration rates using undersampling patterns, such as pseduo-radial, carestian, etc., simulating MRI acquisition under accelerated conditions. This paired dataset, consisting of undersampled LF and fully sampled high-field k-space, enables the network to learn the mapping required for accurate frequency-domain reconstruction. Additionally, the model enhances image quality by recovering missing frequency components lost during undersampling. Once trained, the model is capable of reconstructing high-field-like MR images from undersampled LF k-space inputs. To quantify uncertainties in the reconstruction process, we implement an ensemble approach using cross-validation as in \cite{lakshminarayanan2017simple, dutschmann2023large}. For each test input, the ensemble mean yields the final reconstruction, while the pixel-wise standard deviation across predictions provides an uncertainty map reflecting predictive variability under undersampling and OOD shift.}

% \vspace{-0.2cm}
\subsection{Deep learning architecture configuration}
\label{ssec:arch}
% \vspace{-0.15cm}
Figure \ref{fig:U-Net} illustrates a modified three-layer U-Net designed for reconstructing complex-valued MRI k-space, represented as two real-valued channels for the real and imaginary components. We use standard convolutional layers to maintain a consistent baseline for comparing models trained in the spatial and k-space domains. The network maps this two-channel input to a high-field-like k-space output in the same format. The encoder section of the network comprises three convolutional blocks. The first block includes two consecutive $3 \times 3$ convolutional layers with 64 output channels each, followed by a $2 \times 2$ max-pooling layer. The second block contains two $3 \times 3$ convolutional layers with 128 output channels each, followed by another max-pooling layer. The third block consists of two $3 \times 3$ convolutional layers with 256 output channels each, followed by max-pooling. At the bottleneck of the architecture, two $3 \times 3$ convolutional layers with 512 output channels capture the most abstract representations of the input data. The decoder mirrors the encoder, consisting of three upsampling stages. The first stage employs a $2 \times 2$ transposed convolution to reduce the channel depth from 512 to 256, followed by two consecutive $3 \times 3$ convolutional layers with 256 output channels, and then concatenated with the corresponding encoder block through skip connections. The second stage upsamples from 256 to 128 channels, followed by two consecutive $3 \times 3$ convolutional layers with 128 output channels, which are concatenated with the second encoder block. The third stage upsamples from 128 to 64 channels, followed by two consecutive $3 \times 3$ convolutional layers with 64 output channels, which are then concatenated with the first encoder block. All convolutional layers utilise ReLU activation functions, except for the final $1 \times 1$ convolutional layer, which outputs two channels representing the reconstructed real and imaginary k-space components. This three-block dual-channel U-Net was chosen after internal model selection on the validation split within 3-fold cross-validation, and was used for all reported experiments.
% This configuration was selected following systematic evaluation of architectural variants and hyperparameter settings, with a three-block dual-channel U-Net combined with optimised hyperparameters and 3-fold cross-validation consistently yielding the strongest reconstruction performance.

% \vspace{-0.01cm}
\section{Experiments and Results}
% \vspace{-0.01cm}
\subsection{Datasets}
\label{ssec:data}
\begin{figure}[ht]
\centering
\includegraphics[width=0.995\columnwidth]{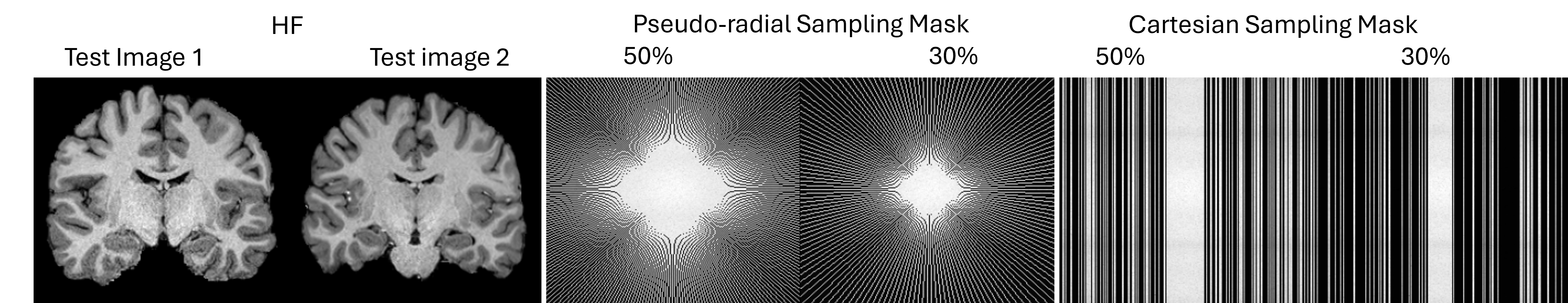}
% \vspace{-0.4cm}
\caption{High-field (HF) MR test images, and example pseudo-radial as well as Cartesian undersampling binary masks at 50\% and 30\% sampling rates.} 
\label{fig:PRS_HF(OODs)}
% \vspace{-0.5cm}
\end{figure}

% \vspace{-0.05cm}
To evaluate the performance of our proposed model, we use high-resolution T1-weighted (T1w) volumes from the WU-Minn Human Connectome Project (HCP), acquired at 3T with $0.7$ mm isotropic voxels \cite{Sotiropoulos2013Oct}. From this dataset, we randomly select 40 subjects (equivalent to 4000 slices) for training and validation, while reserving 10 subjects (equivalent to 200 slices) for testing. LF-MR images are synthesised from the high-field MR counterparts using the stochastic low-field simulator in \cite{lin2019deep,lin2023low, eldaly2024alternative}, which models tissue-specific contrast changes to replicate LF SNR and contrast degradation. The InD dataset used for training is synthesised with parameters constrained by a Mahalanobis distance $< 1$, ensuring that the SNR in white matter (WM) is higher than in gray matter (GM) to maintain tissue contrast compatible with T1-weighted images. On the other hand, the OOD test dataset is synthesised by sampling SNR parameters from a distribution reflecting ultra-low field T1-weighted MRI images, introducing significant deviations from the Gaussian distribution used for the InD data for training. To simulate accelerated acquisition, LF k-space data are undersampled using pseudo-radial and Cartesian sampling patterns at 50\% and 30\% sampling rates, alongside their fully sampled (100\%) references, before feeding the data into the network. The resulting reconstructions are compared across these acceleration levels in both spatial and k-space image quality transfer settings (referred to as ``kIQT'' and ``sIQT''), to assess how each domain handles undersampling artifacts. Figure \ref{fig:PRS_HF(OODs)} shows examples of two high-field test images alongside the corresponding pseudo-radial and Cartesian sampling masks at 50\% and 30\% sampling rates, respectively.

% \subsection{Deep learning architecture configuration}
% \label{ssec:arch}
% \vspace{-0.05cm}
\subsection{Model Training and Testing}
\label{ssec:training}
% \vspace{-0.05cm}
{We perform 3-fold cross-validation using 4000 training slices, with each fold using one-third of the slices for validation and the remaining two-thirds for training on our proposed network. For model optimisation during training, we use the Adam optimiser with a learning rate of $10^{-3}$ and weight decay of $10^{-6}$. The model is trained with a batch size of 8, using a combined loss function that incorporates both mean absolute error (MAE) and mean squared error (MSE). This hybrid loss function strikes a balance between robustness and fidelity in the reconstructions. The training process is conducted in PyTorch on an NVIDIA A100 GPU (80GB), with each fold consisting of 150 epochs. The best model is selected based on the minimum validation loss, ensuring that the most accurate and reliable model is used for final OOD testing. Notably, the training process takes approximately 4 hours, while inference is completed in under 1 minute, demonstrating the model's clinical feasibility. OOD LF-MRI slices are reconstructed at various sampling rates, and the results are compared to high-field references using peak signal-to-noise ratio (PSNR) and structural similarity index measure (SSIM). The mean and standard deviation of these values are reported across the folds and test slices. Prior to running the final OOD evaluation, we performed internal model selection using the validation split within the same 3-fold cross-validation protocol. We compared a small set of architectural depths and training hyperparameter settings, and selected the configuration that achieved the lowest validation loss while maintaining practical inference cost. The resulting three-block dual-channel U-Net and the training settings reported above were then fixed and used consistently across folds for the final OOD testing and metric reporting.}
% \vspace{-0.05cm}
\subsection{Quantitative analysis}
% \vspace{-0.05cm}
{The quantitative evaluation of our proposed approach on OOD data is summarised in Table \ref{tab:PRS_CARiqt}, which presents PSNR and SSIM metrics for image quality enhancement using undersampled k-space data at different undersampling patterns (pseudo-radial and Cartesian patterns) and different sampling ratios. This is compared to spatial-domain IQT (sIQT) \cite{lin2023low}. As anticipated, the baseline LF images demonstrate poor SSIM and lower PSNR across both domains, with performance deteriorating as the undersampling rate increases. For instance, SSIM decreases from 0.7466 at full sampling to 0.2606 at 30\% pseudo-radial sampling, indicating significant structural loss under heavy undersampling. In contrast, both IQT approaches deliver substantial improvements over the low-field baseline, validating the effectiveness of IQT in recovering high-field-like results. Specifically, sIQT boosts SSIM to 0.8077 at 30\% pseudo-radial sampling, compared to 0.2606 for the raw low-field data. When comparing the two IQT strategies, kIQT consistently achieves higher SSIM and PSNR values than sIQT across all undersampling rates and patterns. Notably, this performance improvement is also observed in the Cartesian sampling pattern outcomes. These results, especially when applied to undersampled k-space, demonstrate the potential for significantly accelerating MRI acquisition without compromising reconstruction quality. This highlights the advantage of directly leveraging k-space information for IQT, rather than relying on post-processing steps applied after spatial domain reconstruction.}

% \vspace{-0.1cm}
\subsection{Qualitative analysis}
\begin{figure}[ht]
\centering
\includegraphics[width=0.995\columnwidth]{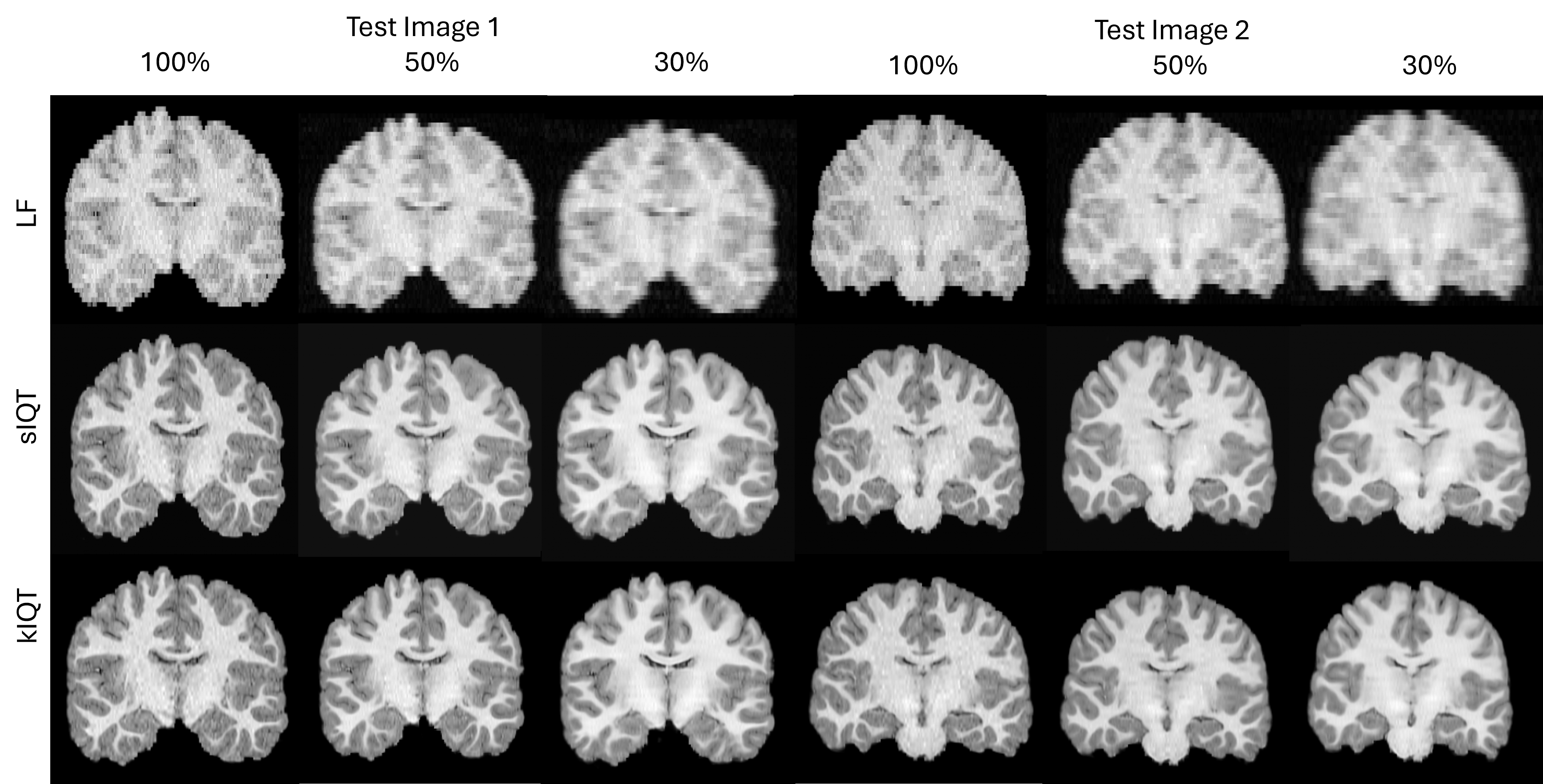}
% \vspace{-0.4cm}
\caption{Reconstruction results of the LF images, and IQT in the spatial and k-space domains, at different Cartesian under-sampling rates (100\%, 50\%, 30\%) using two test images.} 
\label{fig:CAR_IQT(OODs)}
% \vspace{-0.5cm}
\end{figure}

\begin{figure}[ht]
\centering
\includegraphics[width=0.995\columnwidth]{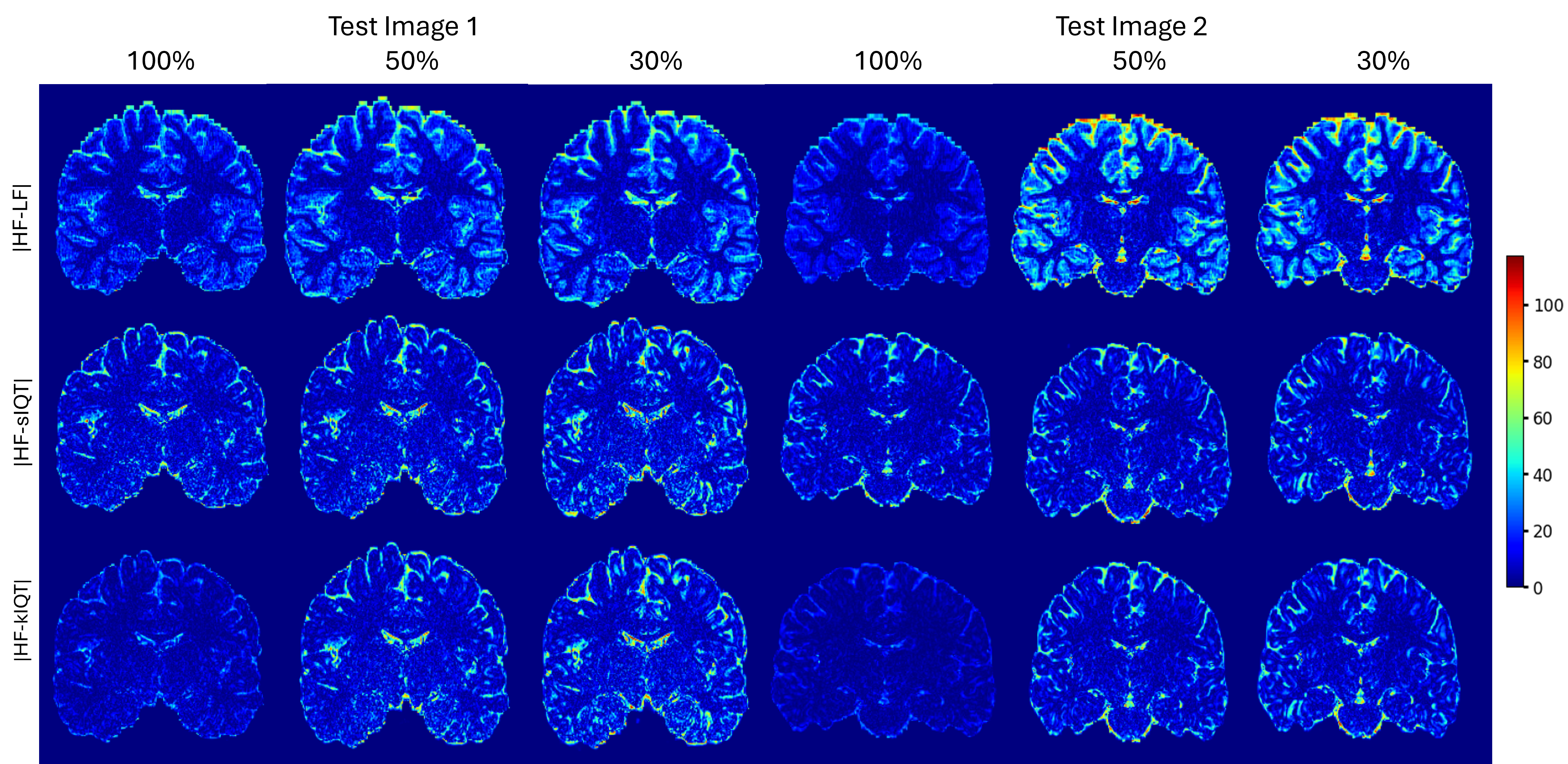}
% \vspace{-0.4cm}
\caption{Absolute error maps between high-field images in Figure \ref{fig:PRS_HF(OODs)}, and the reconstructions in Figure \ref{fig:CAR_IQT(OODs)} at different Cartesian under-sampling rates - $|\text{HF} - \text{LF}|$ in row 1, $|\text{HF} - \text{IQT-spatial}|$ in row 2 and $|\text{HF} - \text{IQT-k-space}|$ in row 3.} 
\label{fig:CAR_EM(OODs)}
% \vspace{-0.3cm}
\end{figure}

\begin{figure}[ht]
\centering
\includegraphics[width=0.995\columnwidth]{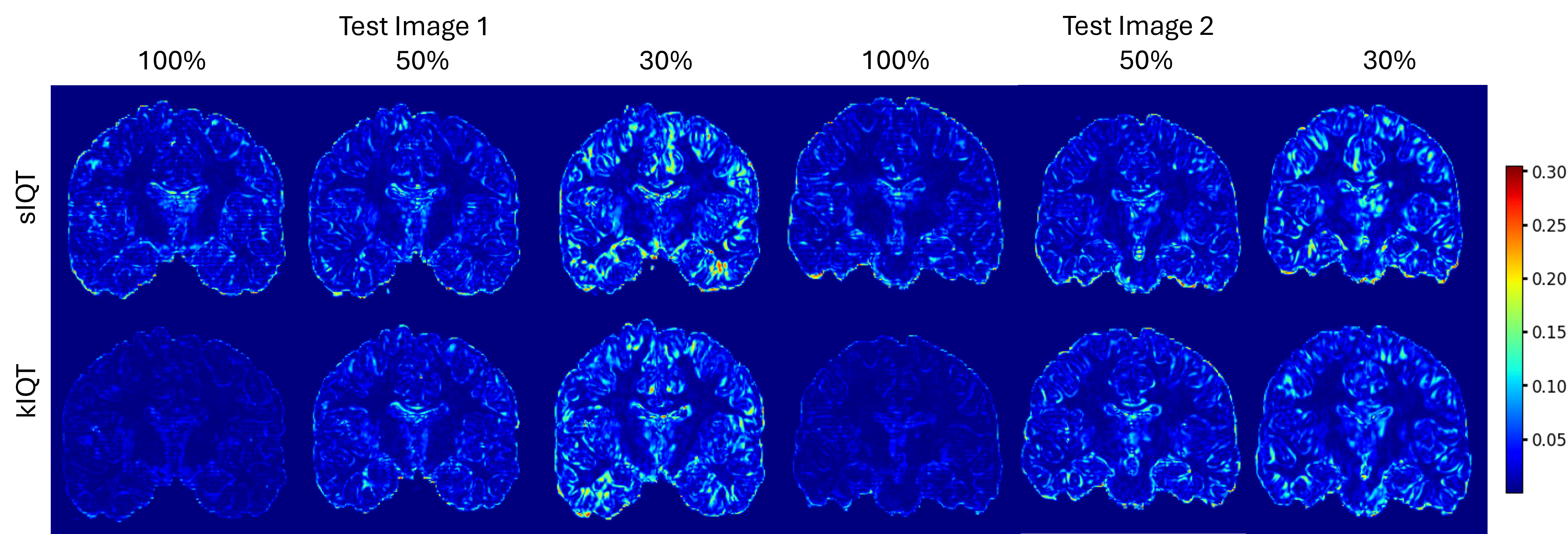}
% \vspace{-0.4cm}
\caption{Uncertainty maps at different Cartesian under-sampling rates across two test images in the spatial and k-space domains.} 
\label{fig:CAR_UQ(OODs)}
% \vspace{-0.5cm}
\end{figure}

% \vspace{-0.1cm}
To further evaluate the performance of the proposed method, Figure \ref{fig:CAR_IQT(OODs)} presents reconstructed brain images from the OOD test dataset, obtained using k-space and spatial-domain reconstruction approaches across Cartesian sampling patterns at various undersampling ratios. The behavior using the Pseudo-radial sampling is similar and therefore is not included here. Figure \ref{fig:CAR_EM(OODs)} presents the corresponding error maps, highlighting the absolute differences between the original high-field images in Figure \ref{fig:PRS_HF(OODs)} and their LF and IQT reconstructions in Figure \ref{fig:CAR_IQT(OODs)}. These qualitative observations are consistent with the quantitative results reported in Table \ref{tab:PRS_CARiqt}. As anticipated, LF reconstructions degrade noticeably at lower sampling rates. This effect is particularly pronounced at 30\% sampling, where structural distortions and intensity inconsistencies are more evident. The error maps confirm these issues, with higher residuals concentrated around anatomical edges, indicating regions of substantial reconstruction error. In contrast, IQT reconstructions more effectively preserve fine details, producing images with improved structural fidelity and reduced artifacts. Even at 30\% sampling, IQT maintains clearer anatomical delineation with minimal blurring. The associated error maps support this improvement, showing substantially lower residuals relative to LF reconstructions. A comparison between k-space and spatial-domain reconstructions reveals that differences become increasingly noticeable at reduced sampling rates (50\% and 30\%). Additionally, the effect of sample reduction on model confidence was evaluated through UQ, as illustrated in Figure \ref{fig:CAR_UQ(OODs)}. Generally, confidence decreases as the sampling rate is reduced. The IQT uncertainty maps highlight regions of higher uncertainty, predominantly near structural boundaries, where reconstruction errors are more likely to occur in unseen, real-world conditions. In this context, elevated uncertainty should be interpreted as reduced model confidence in the reconstruction. These areas roughly correspond to regions with larger residuals in the $|\text{HF} - \text{LF}|$ and $|\text{HF} - \text{IQT}|$ error maps, suggesting that the uncertainty estimates capture potential ambiguities in the OOD reconstructions.

\begin{table}[ht]
\caption{Average and standard deviation PSNR and SSIM measures for k-space and spatial domain IQT at different pseudo-radial and Cartesian sampling rates using OOD data.}
\centering
\footnotesize
\setlength{\tabcolsep}{3pt}
\renewcommand{\arraystretch}{1.1}
\resizebox{\linewidth}{!}{%
\begin{tabular}{l|l|c|c|c}
\hline
\textbf{Domain} & \textbf{Metric} & \textbf{100\%} & \textbf{50\%} & \textbf{30\%} \\
\hline\hline
\multicolumn{5}{c}{\textbf{Pseudo-radial Sampling}} \\
\hline\hline
\textbf{Low-Field (LF)} 
& SSIM & 0.7466 $\pm$ 0.0495 & 0.3070 $\pm$ 0.0727 & 0.2606 $\pm$ 0.0565 \\
& PSNR & 21.80 $\pm$ 1.9568 & 17.76 $\pm$ 0.6992 & 17.72 $\pm$ 0.7950 \\
\hline\hline
\textbf{sIQT} 
& SSIM & 0.8163 $\pm$ 0.0496 & 0.8081 $\pm$ 0.0539 & 0.8077 $\pm$ 0.0587 \\
& PSNR & 24.74 $\pm$ 1.6038 & 24.01 $\pm$ 2.071 & 23.64 $\pm$ 2.5127 \\
\hline\hline
\textbf{kIQT} 
& SSIM & \textbf{0.8219 $\pm$ 0.0550} & \textbf{0.8159 $\pm$ 0.0675} & \textbf{0.8138 $\pm$ 0.0725} \\
& PSNR & \textbf{25.99 $\pm$ 2.5793} & \textbf{25.96 $\pm$ 2.5999} & \textbf{25.95 $\pm$ 2.5547} \\
\hline\hline
\multicolumn{5}{c}{\textbf{Cartesian Sampling}} \\
\hline\hline
\textbf{Low-Field (LF)} 
& SSIM & 0.7466 $\pm$ 0.0495 & 0.4733 $\pm$ 0.0422 & 0.4513 $\pm$ 0.0343 \\
& PSNR & 21.80 $\pm$ 1.9568 & 17.62 $\pm$ 0.5397 & 17.61 $\pm$ 0.5285 \\
\hline\hline
\textbf{sIQT} 
& SSIM & 0.8163 $\pm$ 0.0496 & 0.8048 $\pm$ 0.0515 & 0.7859 $\pm$ 0.0479 \\
& PSNR & 24.74 $\pm$ 1.6038 & 23.95 $\pm$ 2.0854 & 23.16 $\pm$ 1.4648 \\
\hline\hline
\textbf{kIQT} 
& SSIM & \textbf{0.8219 $\pm$ 0.0550} & \textbf{0.8160 $\pm$ 0.0566} & \textbf{0.7992 $\pm$ 0.0547} \\
& PSNR & \textbf{25.99 $\pm$ 2.5793} & \textbf{25.92 $\pm$ 2.5516} & \textbf{25.56 $\pm$ 2.4367} \\
\hline
\end{tabular}}
% \vspace{-0.4em}
\label{tab:PRS_CARiqt}
\end{table}
%%%%%%%%%%  


\subsection{Comparison with existing work}
While the main focus of this work is to reconstruct high-field-like MR images from under-sampled LF-MRI k-space and to compare this approach with spatial-domain image quality transfer/super-resolution, we also evaluate the performance of our method by comparing it to several existing techniques. 
% In particular, we benchmark our approach against deep dictionary learning IQT (IQT-DDL) \cite{eldaly2024alternative} and conditional flow matching IQT (IQT-CFM) methods \cite{Tien2026ISBI}, in addition to nearest-neighbor interpolation and the counterpart spatial domain IQT of the proposed approach. 
In particular, we benchmark our approach against deep dictionary learning IQT (IQT-DDL) \cite{eldaly2024alternative}, nearest-neighbor interpolation, and the counterpart spatial domain IQT of the proposed approach. Since all the comparison methods operate in the spatial domain, they were tested using the same input data and assessed against super-resolution applied to fully sampled (100\%) reconstructions. Table \ref{tab:EXT_METiqt} presents the PSNR and SSIM values for all methods using the OOD dataset. The results show that the proposed kIQT approach consistently achieves higher PSNR and SSIM scores, outperforming all baseline methods. This demonstrates that kIQT produces more reliable and higher-quality reconstructions than existing spatial-domain techniques.
% \vspace{0.4cm}

%%%%%

% % \setlength{\arrayrulewidth}{0.4pt}
% \setlength{\tabcolsep}{5pt}
% \renewcommand{\arraystretch}{0.99} 
\begin{table}
\caption{Comparison with other existing methods (in the spatial domain) using 100\% sampling.}
\centering
% \resizebox{\columnwidth}{!}{%
% \tiny
% \scriptsize
\begin{tabular}{ |c|c|c| }
\hline
\textbf{Methods} & \textbf{Metric} & \textbf{OOD} \\
\hline

Low-Field (LF)  & SSIM  & 0.7466 $\pm$ 0.0495 \\
                & PSNR  & 21.80 $\pm$ 1.9568 \\
\hline

IQT-DDL        & SSIM  & 0.7660 $\pm$ 0.0610 \\
               & PSNR  & 24.25 $\pm$ 0.823 \\
\hline

% IQT-CFM        & SSIM  & 0.7700 $\pm$ 0.0500 \\
%                & PSNR  & 24.33 $\pm$ 0.8210 \\
% \hline

sIQT           & SSIM  & 0.8163 $\pm$ 0.0496 \\
               & PSNR  & 24.74 $\pm$ 1.6038 \\
\hline

\textbf{kIQT}  & SSIM  & \textbf{0.8219 $\pm$ 0.0550} \\
               & PSNR  & \textbf{25.99 $\pm$ 2.5793} \\
\hline
\end{tabular}
% }
\label{tab:EXT_METiqt}
% \vspace{-0.3cm}
\end{table}

%%%%%

% \vspace{-0.25cm}
\section{Conclusion}
% \vspace{-0.25cm}
In this work, we introduced a novel framework for image quality transfer that simultaneously reconstructs high-field magentic resonance like images from undersampled k-space data and quantifies the effect of reduced sampling. Unlike traditional methods, which treat image reconstruction and quality enhancement as separate stages, our approach directly leverages k-space information which allows for accelerated scan times through k-space undersampling while maintaining diagnostic image quality. Experiments on an out-of-distribution low-field brain MRI dataset demonstrated that the proposed k-space–driven image quality transfer approach outperforms spatial-domain counterparts and achieves image quality comparable to fully sampled data, despite substantial undersampling. Moreover, the integration of uncertainty quantification provides valuable insight into the reliability of the reconstructed images. In future work, we plan to explore adaptive k-space sampling strategies driven by uncertainty maps, optimising acquisition efficiency while ensuring high-quality reconstructions from minimal data.

\bibliographystyle{IEEEbib}
\bibliography{strings,refs}

\end{document}